\documentclass[conference]{IEEEtran}
\IEEEoverridecommandlockouts
\usepackage[T1]{fontenc}
\makeatletter
\let\NAT@parse\undefined
\makeatother

\usepackage[numbers,sort&compress]{natbib}

\usepackage{balance}
\usepackage{amsmath,amssymb,amsfonts,graphicx,booktabs,array,xcolor,url}  
\usepackage{tabularx}
\usepackage{algorithm,algpseudocode,tikz}
\usetikzlibrary{arrows.meta,positioning}
\usepackage[hidelinks]{hyperref}
\usepackage{colortbl}
\usepackage{multirow}

\providecommand{\abyes}{%
  \textcolor{green!50!black}{\ensuremath{\checkmark}}%
}
\providecommand{\abno}{%
  \textcolor{red!75!black}{\ensuremath{\times}}%
}
\makeatletter
\let\xmind@originalmakecaption\@makecaption
\def\xmind@tabletype{table}

\long\def\@makecaption#1#2{%
  \ifx\@captype\xmind@tabletype
    \noindent
    \parbox{\linewidth}{%
      \normalfont\small
      \noindent #1: #2\par
    }%
    \par\vspace{4pt}%
  \else
    \xmind@originalmakecaption{#1}{#2}%
  \fi
}
\makeatother
\definecolor{settingblue}{HTML}{EAF2F8}
\newcommand{\settingrow}{\rowcolor{settingblue}}
\definecolor{draftred}{RGB}{165,35,40}
\definecolor{inkblue}{RGB}{34,73,99}
\definecolor{paleblue}{RGB}{235,242,247}

\newcommand{\xmind}{XMind}

\newcommand{\act}[1]{\texttt{#1}}
\begin{document}
\title{\LARGE\bfseries XMind: Thinking Twice via Cross-Memory Reasoning for LLM-Based Autonomous Driving}
\author{Zuyi Guo$^{1}$, Xianjie Dai$^{1}$, Lei Yang$^{2}$, Xiaocong Zhao$^{3}$, Jingguang Li$^{1}$, Kitae Jang$^{1}$, and Heye Huang$^{1,\dagger}$%
\thanks{$^{\dagger}$Corresponding author: {\tt\small heye.huang@kaist.ac.kr}.}%
\\[1.5ex]
$^{1}$Korea Advanced Institute of Science and Technology \\
$^{2}$Nanyang Technological University \\ 
$^{3}$Tongji University 
}
\maketitle
\thispagestyle{empty}
\pagestyle{empty}

\begin{abstract}
Memory-augmented large language model (LLM) driving agents can reuse past experience, yet previously learned corrections often fail to transfer when traffic conditions change. A key limitation is that existing memories mainly record observed outcomes, providing little evidence about alternative actions. We introduce XMind, a training-free framework that enables LLM-based driving to ``Think Twice'' through cross-memory reasoning. XMind integrates working memory of recent traffic states with episodic memory of critical risk events. The agent first makes an initial decision from the current context and then cross-checks it against retrieved historical experience. If a conflict arises, it makes a second decision guided by the retrieved evidence. To construct reliable episodic memory, XMind evaluates alternative actions through counterfactual rollouts under perturbed traffic conditions and stores the verified outcomes as structured risk capsules. These capsules are retrieved by jointly matching spatial risk patterns and scenario semantics. Across closed-loop highway simulations, XMind achieves success rates of 83.3–93.3\%, exceeding the strong baseline by 6.6 percentage points on average. Evaluation on K-Risk further demonstrates lower decision risk on unseen real-world high-risk interactions. Ablation studies confirm the complementary benefits of working memory, risk--text retrieval, and counterfactual verification. Code is available at  \url{https://github.com/xiangxiangliuwo/XMind}.
\end{abstract}


\section{Introduction}
\label{sec:intro}

Recent advances have substantially improved autonomous driving performance in normal traffic scenarios, yet reliable decision-making in long-tail scenarios remains challenging. Because events such as abrupt braking, cut-ins, and dense multi-vehicle interactions are rare, the experience gained from such events can be particularly valuable. For large language model (LLM) driving agents, external memory provides a way to reuse such experience without updating model parameters~\citep{wen2024dilu, shinn2023reflexion}. Reliable experience reuse, however, raises two questions: \emph{what should be remembered from a past risk event, and when should that memory influence the current decision?}

Existing LLM driving agents retrieve past experience and reflections to support decision-making~\citep{wen2024dilu, mao2023language, jiang2024koma}, while RESPOND~\citep{chen2026respond} further organizes memories using structured risk patterns. However, similar scenarios or risk layouts do not necessarily require the same maneuver. More importantly, existing memories mainly record the outcome of the executed action, providing little evidence about alternative actions. Counterfactual memories such as CounterMem~\citep{pu2026countermem} verify corrective actions and their applicability, but retain little information about the other actions available at the same critical state. Safety-critical driving therefore requires multi-action evidence that distinguishes recoverable, dangerous, and uncertain alternatives under changing traffic conditions.

A second challenge is deciding when retrieved experience should affect online reasoning. Existing memory-augmented agents typically use retrieved experience whenever a match is found, allowing weakly related memory to interfere with otherwise reasonable decisions. This motivates a selective mechanism in which memory is used only when it provides evidence that challenges the agent's initial choice.

\begin{figure}[!t]
    \centering
    \includegraphics[width=\columnwidth]{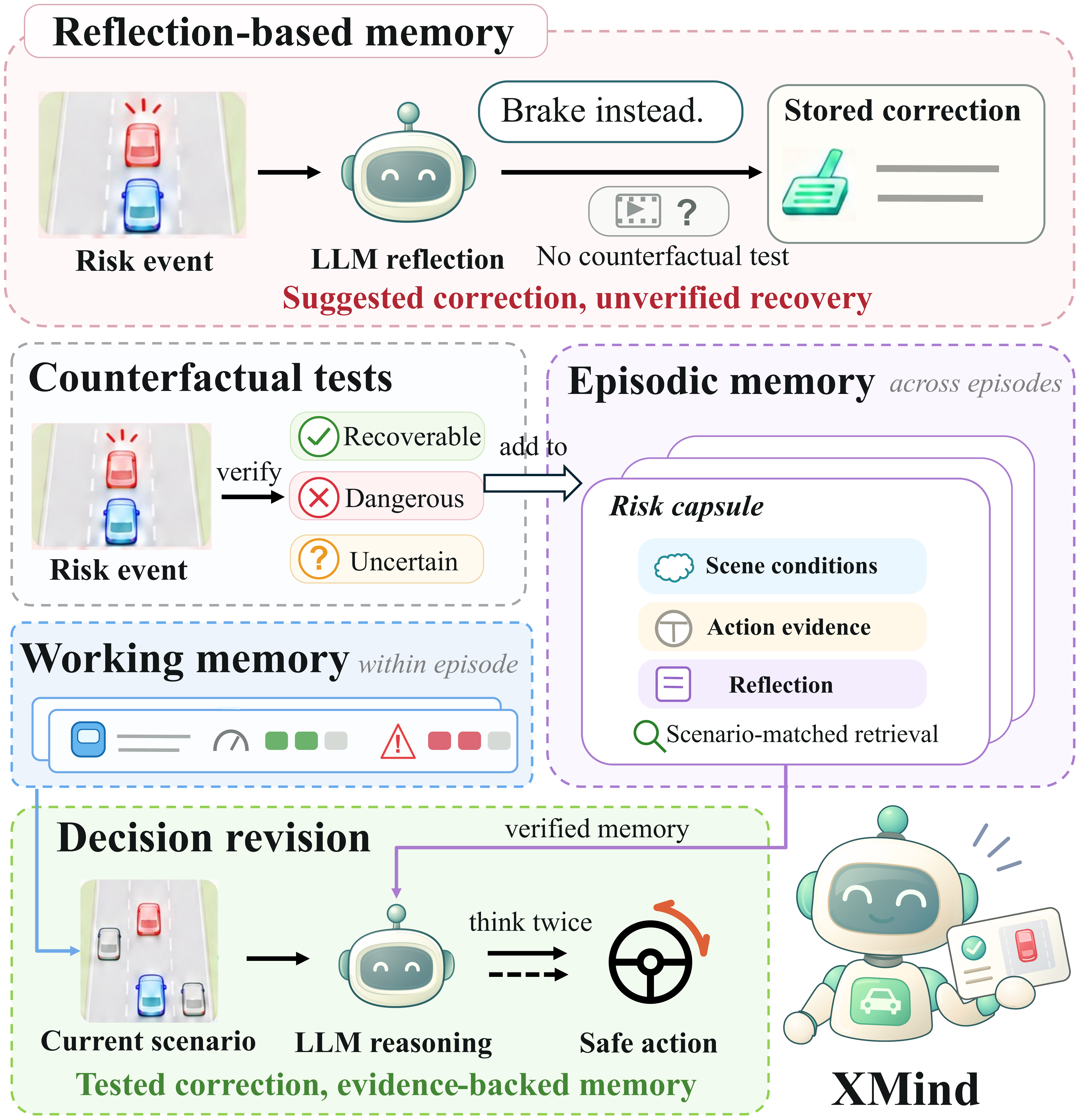}
    \caption{From reflection-based correction to ``Think Twice'' reasoning.
    Top: LLM reflection suggests a correction without
    counterfactual verification of recoverability.
    Bottom: \xmind{} tests candidate actions under traffic
    perturbations and continuation policies, stores the resulting
    evidence in episodic risk capsules, and triggers ``Think Twice''
    cross-memory reasoning when retrieved evidence disagrees with
    the initial decision.}
    \label{fig:concept}
\end{figure}

To address these challenges, we propose \xmind{}, a training-free cross-memory reasoning framework for LLM-based autonomous driving. \xmind{} combines short-term working memory of recent traffic states with long-term episodic memory of critical risk events, as illustrated in Fig.~\ref{fig:concept}. In the offline phase, candidate actions at critical states are evaluated through counterfactual rollouts under traffic perturbations and stored as structured risk capsules. During online decision-making, the LLM first makes an initial decision from working memory and then cross-checks it against retrieved episodic evidence. When the two disagree, \xmind{} triggers a second, evidence-guided decision, allowing the system to ``Think Twice'' while keeping the underlying LLM frozen.

The primary contributions of this paper are as follows:
\begin{itemize}
    \item We propose \xmind{}, a cross-memory reasoning framework that lets an LLM driving agent ``Think Twice'': its initial decision from working memory is cross-checked against episodic risk memory, and disagreement triggers an evidence-guided revision.
    \item We construct a counterfactually verified episodic risk memory, whose risk capsules record recoverable, dangerous, and uncertain actions at decision-critical states and are retrieved through joint risk--text matching.
    \item We validate \xmind{} through closed-loop simulations, open-loop evaluation on K-Risk~\citep{huang2026knowledge}, and ablation studies, showing higher success rates than strong baselines and reduced risk on recorded high-risk events.
\end{itemize}

\section{Related Work}
\label{sec:related}
\subsection{LLM-based Driving Agents}

LLM driving frameworks use different representations and retrieval mechanisms to reuse past experience. DiLu~\citep{wen2024dilu} retrieves prior reasoning traces and actions from scenario descriptions, while Agent-Driver~\citep{mao2023language} separates commonsense and driving memories. RAG-Driver~\citep{yuan2024rag} learns a shared retrieval space for visual and control features, LeapVAD~\citep{ma2025leapvad} uses action-aware contrastive representations, and LiloDriver~\citep{yao2025lilodriver} incrementally associates scenarios with planning strategies for context-aware decision support. In contrast, \xmind{} combines working memory with episodic risk memory and uses counterfactually verified evidence to revise decisions only when retrieved experience conflicts with the initial choice.

Risk information can also be incorporated into the retrieval space. Driver risk field (DRF)~\citep{kolekar2020human} estimates perceived risk from a state-dependent spatial field combined with environmental costs. Driving safety field captures coupled driver--vehicle--road interactions~\citep{wang2021towards} and yields dynamic risk maps when combined with the intentions of surrounding vehicles~\citep{huang2020probabilistic}. RiskNet~\citep{liu2026risknet} couples interaction fields with probabilistic behavior prediction to forecast risk in long-tail scenarios. SafeDrive~\citep{zhou2026safedrive} extends DRF to quantify coupled multi-agent risks for LLM driving reasoning, while RESPOND~\citep{chen2026respond} encodes ego-centric spatial risk in a discrete matrix and retrieves experience through exact pattern matching.

\subsection{Experience Memory and Reflection}

Memory systems for language agents differ in the knowledge they store and the signals used to update it. ReasoningBank~\citep{ouyang2026reasoningbank} uses LLM assessments of trajectory outcomes to extract structured strategy entries from both successes and failures. EvolveR~\citep{wu2025evolver} maintains a principle repository through semantic deduplication and utility scoring, and combines experience reuse with reinforcement learning policy updates. ECHO~\citep{hu2025sample} rewrites trajectories in hindsight and compresses them into reusable experience.

Driving memory also varies in its temporal scope and role in decision making. AD-Memo~\citep{yan2026vision} retains information about objects and events within a driving episode to support subsequent decisions. In this work, \xmind{} focuses on building knowledge for handling risks across episodes. It stores action outcomes and supporting rollout evidence at specific decision states, allowing the agent to consult the evidence behind past experience when making subsequent decisions.

\subsection{Counterfactual Evaluation for Driving Decisions}

Evaluating driving decisions in high-risk scenarios requires assessing not only whether collisions can be avoided, but also how alternative actions affect future outcomes. Nishiyama et al.~\citep{nishiyama2020discovering} use counterfactual simulation to distinguish avoidable from unavoidable planner failures. FREA~\citep{chen2024frea} approximates the ego vehicle's feasible region and constrains adversarial background behavior, while WoTE~\citep{li2025end} predicts action-conditioned futures with a BEV world model and evaluates candidate trajectories using a reward model. Recent work further studies counterfactual prediction in driving world models, highlighting the difficulty of estimating futures under alternative ego actions~\citep{zhang2026can}.

Counterfactual evaluation can also provide supervision and reusable experience. CF-VLA~\citep{peng2026counterfactual} constructs counterfactual correction data through rollout, filtering, and labeling, while C-CoT~\citep{tian2026c} uses a meta-action evaluation tree to supervise safety reasoning. These methods mainly use counterfactual outcomes for training or correction. In contrast, \xmind{} stores multi-action evaluation results as persistent episodic risk memory, allowing verified evidence to be retrieved across episodes and used to reconsider subsequent driving decisions.

\section{Method}
\label{sec:method}

\begin{figure*}[t]
    \centering
    \includegraphics[width=\textwidth]
    {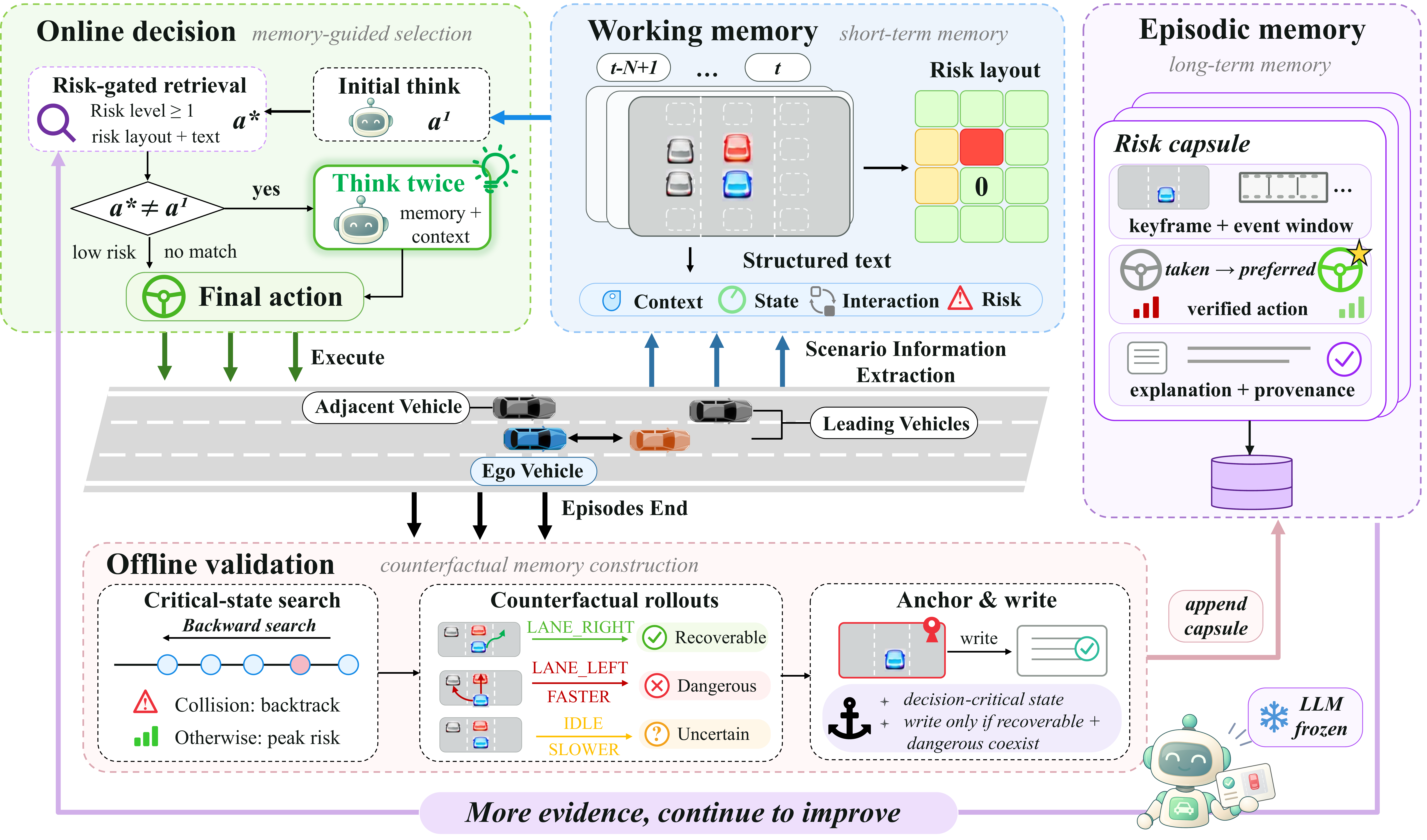}
    \caption{Overview of \xmind{}. Working memory provides recent traffic context and a spatial risk layout. Online decision-making uses risk-gated episodic retrieval to cross-check the initial decision, triggering ``Think Twice'' reasoning when the retrieved preferred action differs from the initial choice. Offline validation evaluates candidate actions at critical states through counterfactual rollouts, distinguishing recoverable, dangerous, and uncertain actions. The resulting evidence is stored as risk capsules in episodic memory for reuse across episodes.
    The LLM parameters remain frozen throughout the whole process.}
    \label{fig:framework}
\end{figure*}

\xmind{} comprises working memory of recent traffic states and episodic memory of counterfactually verified risk experience accumulated across episodes. As shown in Fig.~\ref{fig:framework}, the offline phase builds risk capsules through counterfactual verification, while the online phase combines both memories for cross-memory reasoning. The LLM remains frozen throughout.

\subsection{Risk Representation and Working Memory}
\label{subsec:risk_wm}

\xmind{} makes decisions at the meta-action level. At decision step $t$, it selects $a_t \in \mathcal{A}$, where $\mathcal{A}$ denotes the meta-action space for \xmind{}. A low-level controller converts $a_t$ into vehicle motion.

The scenario encoder organizes the current observation into a structured frame:
\begin{equation}
f_t = \left(t,c,s_t^e,\mathcal{V}_t\right),
\label{eq:structured_frame}
\end{equation}
where $c$ denotes the scenario type, $s_t^e$ is the ego state (identifier, position, longitudinal and lateral velocities, acceleration, and lane), and $\mathcal{V}_t$ denotes the surrounding vehicles.

Surrounding vehicles are mapped to 11 ego-centric directional slots,
$\mathcal{V}_t=(v_t^d)_{d\in\mathcal{D}}$, including eight nearby
directions and three additional forward slots.
Each occupied slot stores
$v_t^d=(s_t^j,r_{t,j},\ell_{t,j})$,
where $s_t^j$ is the vehicle state, $r_{t,j}\in[0,1]$ is its interaction
risk with the ego vehicle, and $\ell_{t,j}$ is the corresponding risk level.
Each slot retains the nearest relevant vehicle, and unoccupied slots are represented as \texttt{null}.

We quantify pairwise interaction risk using DRF~\citep{kolekar2020human}.
Unlike RESPOND~\citep{chen2026respond}, which uses normalized geometric
overlap, we measure the fraction of the reference risk field contained in
the overlap region:
\begin{equation}
r_{t,j}^{\mathrm{DRF}}
=
\frac{
\sum_{p\in\Omega_e\cap\Omega_j}F_{\mathrm{ref}}(p)
}{
\sum_{p\in\Omega_{\mathrm{ref}}}F_{\mathrm{ref}}(p)
}.
\label{eq:pairwise_drf}
\end{equation}
Here, $F_{\mathrm{ref}}$ is the risk field of the trailing vehicle, $\Omega_{\mathrm{ref}}\in\{\Omega_e,\Omega_j\}$ is its field region, and $\Omega_e$ and $\Omega_j$ denote the field regions of the ego vehicle and vehicle $j$, respectively.

For longitudinal closing interactions, we additionally use time to collision
(TTC)~\cite{hayward1972near} to quantify the risk:
\begin{equation}
r_{t,j}^{\mathrm{TTC}}
=
\begin{cases}
1, & g\leq0,\\
0, & g>0,\ \Delta v\leq0,\\
\operatorname{clip}
\left(\dfrac{6-g/\Delta v}{3},0,1\right),
& \text{otherwise},
\end{cases}
\label{eq:ttc_risk}
\end{equation}
where $g$ is the longitudinal gap and $\Delta v$ is the closing speed. For critical longitudinal following scenarios where the lead vehicle’s speed is below \(5\,\mathrm{m/s}\) and the ego vehicle’s closing speed exceeds \(5\,\mathrm{m/s}\), we set $r_{t,j}=r_{t,j}^{\mathrm{TTC}}$. Otherwise, we use $r_{t,j}=r_{t,j}^{\mathrm{DRF}}$.

The risk value is further discretized into three levels:
\begin{equation}
\ell_{t,j}
=
\begin{cases}
0,
& r_{t,j}<0.5,\\
1,
& 0.5\leq r_{t,j}<0.7,\\
2,
& r_{t,j}\geq0.7.
\end{cases}
\label{eq:risk_levels}
\end{equation}
Levels $0$, $1$, and $2$ indicate increasing risk. The continuous value preserves differences in vehicle interactions, and the discrete level is used to construct the spatial risk layout.

The spatial risk layout is denoted by
$\rho_t=\rho(f_t,b_t)\in\{0,1,2\}^{4\times3}$,
where $b_t$ contains the left and right road boundaries read separately from the road network at time $t$. The four rows represent the additional forward positions, the immediate forward positions, lateral positions, and rear positions, while the three columns correspond to the left, ego, and right lanes. Occupied cells contain vehicle risk levels, whereas the ego and empty cells are set to $0$. A missing adjacent lane is assigned level $2$ to represent a lateral road constraint. The layout is used for episodic memory retrieval based on spatial risk similarity.

To preserve recent changes in traffic conditions, consecutive structured frames form the working memory:
\begin{equation}
W_t =
\left(
f_{t-N+1},\ldots,f_t
\right),
\label{eq:working_memory}
\end{equation}
where $W_t$ denotes the working memory at decision step $t$, and $N$ is the number of retained decision frames. The frames are serialized into the decision prompt. Changes in vehicle positions, velocities, accelerations, and risks across these frames provide the model with temporal context for the current interaction.

\subsection{Counterfactual Verification Based on Recoverability}
\label{subsec:counterfactual_verification}
Working memory records recent driving states, while offline counterfactual tests evaluate the recovery options available after different actions. Candidate state selection and memory construction are described in Section~\ref{sec:episodeic_memory}.

At a candidate decision step $\tau$, let $\mathcal A_{\mathrm{av},\tau}\subseteq\mathcal A$ denote the available actions. For each action $a\in\mathcal A_{\mathrm{av},\tau}$, we copy the state saved before action execution and apply traffic condition $\xi_n$. The action is executed for one decision interval, after which a continuation policy $\mu$ completes the rollout.

The policy library $\Pi_{\mathrm{cont}}$ contains IDM car following~\cite{treiber2000congested} and a policy that greedily minimizes risk. All actions share the same $J=4$ traffic conditions $\xi_n$, where $n=1,\ldots,J$. The first condition $\xi_1$ leaves traffic unchanged. The remaining conditions apply multiplicative perturbations to the kinematic features of surrounding vehicles.

Let $z_{\tau}(a,\xi_n,\mu)$ equal one if the ego vehicle collides during the rollout and zero otherwise. An action is recoverable under $\xi_n$ if at least one tested policy avoids collision:
\begin{equation}
\begin{aligned}
\operatorname{Rec}_{\tau}(a,\xi_n)
&=
\mathbf 1\left[
\exists\mu\in\Pi_{\mathrm{cont}}:
z_{\tau}(a,\xi_n,\mu)=0
\right],\\
q_{\tau}(a)
&=
1-\frac{1}{J}\sum_{n=1}^{J}
\operatorname{Rec}_{\tau}(a,\xi_n),
\end{aligned}
\label{eq:counterfactual_recoverability}
\end{equation}
where $\mathbf{1}[\cdot]$ is the indicator function and $q_{\tau}(a)$ is the fraction of tested conditions under which no policy avoids collision. When $q_{\tau}(a)=0$, every tested condition admits at least one collision-free continuation, although the successful policy may differ across conditions.

For a given condition, the set of the policies that avoid collision is defined as:
\begin{equation}
\Pi_{\tau}^{\mathrm{ok}}(a,\xi_n)
=
\left\{
\mu\in\Pi_{\mathrm{cont}}:
z_{\tau}(a,\xi_n,\mu)=0
\right\}.
\label{eq:collision_free_continuations}
\end{equation}

Let $k$ index samples taken at the end of each decision interval, and let $r_{\tau,k}^{\max}(a,\xi_n,\mu)$ denote the maximum directional risk at that sample. When $\Pi_{\tau}^{\mathrm{ok}}(a,\xi_n)\neq\varnothing$, the action risk is defined as the lowest peak risk among collision-free policies:
\begin{equation}
R_{\tau}(a,\xi_n)
=
\min_{\mu\in\Pi_{\tau}^{\mathrm{ok}}(a,\xi_n)}
\max_k r_{\tau,k}^{\max}(a,\xi_n,\mu).
\label{eq:conditional_action_risk}
\end{equation}
If $\Pi_{\tau}^{\mathrm{ok}}(a,\xi_n)=\varnothing$, we set $R_{\tau}(a,\xi_n)=1$.

To compare risk and driving efficiency, we average the action risk and normalized longitudinal progress across traffic conditions:
\begin{equation}
\begin{aligned}
\widehat r_{\tau}(a)
&=
\frac{1}{J}\sum_{n=1}^{J}R_{\tau}(a,\xi_n),\\
p_{\tau}(a)
&=
\frac{1}{J}\sum_{n=1}^{J}
\frac{\Delta x_{\tau}^{e}(a,\xi_n)}
{v_{\mathrm{ref}}T_{\mathrm{cf}}},
\end{aligned}
\label{eq:action_risk_progress}
\end{equation}
where $\Delta x_{\tau}^{e}(a,\xi_n)$ is the ego vehicle's longitudinal displacement from the candidate state to the endpoint of the selected rollout. The reference speed is denoted by $v_{\mathrm{ref}}$, and $T_{\mathrm{cf}}=4\,\mathrm{s}$ is the total rollout duration, including the initial action interval.

The available actions are classified as recoverable, dangerous, or uncertain:
\begin{equation}
\begin{aligned}
\mathcal A_{\tau}^{+}
&=
\{a\in\mathcal A_{\mathrm{av},\tau}:
q_{\tau}(a)=0\},\\
\mathcal A_{\tau}^{-}
&=
\{a\in\mathcal A_{\mathrm{av},\tau}:
q_{\tau}(a)\geq\eta\},\\
\mathcal A_{\tau}^{0}
&=
\mathcal A_{\mathrm{av},\tau}
\setminus
\left(
\mathcal A_{\tau}^{+}\cup\mathcal A_{\tau}^{-}
\right),
\end{aligned}
\qquad \eta=0.5.
\label{eq:recoverability_action_sets}
\end{equation}

When $\mathcal A_{\tau}^{+}$ is nonempty, the preferred action balances progress against risk:
\begin{equation}
a_{\tau}^{*}
=
\underset{a\in\mathcal A_{\tau}^{+}}{\arg\max}
\left[
p_{\tau}(a)-\lambda\widehat r_{\tau}(a)
\right],
\qquad \lambda=0.5.
\label{eq:preferred_action}
\end{equation}

\subsection{Long-term Episodic Memory and Offline Validation}
\label{sec:episodeic_memory}

Episodic memory stores experience around critical decisions. During online execution, we retain snapshots taken before action execution in working memory. A separate snapshot preserves the state at the episode's maximum directional risk. At the end of the episode, the action tests described in Section~\ref{subsec:counterfactual_verification} identify a memory anchor from these states.

For an episode with a collision, let
$\mathcal C_{\mathrm{back}}$ denote the retained decision steps preceding the collision. We examine these steps in reverse order and select the latest one that admits a recoverable action:
\begin{equation}
\tau^{*}
=
\max
\left\{
\tau\in\mathcal C_{\mathrm{back}}:
\mathcal A_{\tau}^{+}\neq\varnothing
\right\}.
\label{eq:memory_anchor}
\end{equation}
If no such step exists in the retained window, no capsule is generated. For an episode without a collision, $\tau^{*}$ is the earliest decision step attaining the episode's maximum directional risk. This state is verified only if its risk reaches level $1$ or above.

The selected state is stored only when it contains both recoverable and dangerous actions:
\begin{equation}
\operatorname{Write}(\tau^{*})
=
\mathbf 1\left[
\mathcal A_{\tau^{*}}^{+}\neq\varnothing
\land
\mathcal A_{\tau^{*}}^{-}\neq\varnothing
\right].
\label{eq:capsule_write_condition}
\end{equation}
This condition retains states in which different actions produce distinct tested outcomes. For a collision episode, $\tau^*$ marks the latest point at which the collision could still be avoided. If no action at $\tau^*$ is dangerous, the action choice there was not decisive under the tested conditions, so the episode yields no capsule; earlier states are not searched because their connection to the collision is weaker.

Each retained event is stored as a structured risk capsule:
\begin{equation}
\begin{aligned}
m_i=\bigl(&
f_{\tau_i^{*}},\,b_i,\,\rho_i,\,
\mathcal H_i,\,\mathcal T_i,\\
&
a_i^{\mathrm{taken}},\,a_i^{*},\,
e_i,\,\mathcal P_i
\bigr),
\end{aligned}
\label{eq:risk_capsule}
\end{equation}
where $\tau_i^{*}$ is the critical decision step, $\mathcal H_i$ and $\mathcal T_i$ store the observed event window and counterfactual action outcomes, respectively. $a_i^{\mathrm{taken}}$ is the executed action, and $a_i^{*}=a_{\tau_i^{*}}^{*}$ is the preferred action from Equation~\eqref{eq:preferred_action}. $e_i$ is a reflection generated by the frozen LLM from $H_i$ and $T_i$, and $P_i$ contains source metadata and verification settings.

The capsule is then appended to persistent episodic memory for retrieval in later episodes:
\begin{equation}
\mathcal M_E^{\,\prime}
=
\mathcal M_E\cup\{m_i\},
\label{eq:episodic_memory_update}
\end{equation}
where $\mathcal M_E$ and $\mathcal M_E^{\,\prime}$ denote the memory before and after insertion, respectively.

\subsection{Memory Retrieval and Online Decision Making}
\label{subsec:online_decision}

The applicability of a stored action outcome depends on the current scenario. Let $\pi_{\theta}$ denote the driving policy implemented by the LLM, with frozen model parameters $\theta$. The prompt $P_t$ contains the system instructions, driving intent, and available actions. The model uses this prompt and working memory $W_t$ to produce an initial action:
\begin{equation}
a_t^{(1)}
=
\pi_{\theta}(P_t,W_t).
\label{eq:initial_online_action}
\end{equation}

\begin{table*}[t]
\centering
\caption{Comparison of driving performance and token usage across highway scenarios with different lane counts and traffic densities. \textbf{Bold} and \underline{underlined} values indicate the best and second best results, respectively. This convention applies to all subsequent tables.}
\label{tab:highway}
\small\setlength{\tabcolsep}{4pt}
\resizebox{\textwidth}{!}{%
\begin{tabular}{l cccc cccc cccc cc}
\toprule
& \multicolumn{4}{c}{4 lanes, density 2.5} & \multicolumn{4}{c}{5 lanes, density 2.5} & \multicolumn{4}{c}{5 lanes, density 3.0} & \multicolumn{2}{c}{Tokens/step}\\
\cmidrule(lr){2-5}\cmidrule(lr){6-9}\cmidrule(lr){10-13}\cmidrule(lr){14-15}
Method & SR$\uparrow$ & SS$\uparrow$ & $\bar v\uparrow$ & $\overline{\mathrm{TTC}}_{\min}$$\uparrow$ & SR$\uparrow$ & SS$\uparrow$ & $\bar v\uparrow$ & $\overline{\mathrm{TTC}}_{\min}$ $\uparrow$ & SR$\uparrow$ & SS$\uparrow$ & $\bar v\uparrow$ & $\overline{\mathrm{TTC}}_{\min}$$\uparrow$ & Input$\downarrow$ & Output$\downarrow$\\
\midrule
DiLu~\citep{wen2024dilu} & 70.0 & 22.13 & \underline{12.99} & 6.33 & 73.3 & 23.27 & \underline{15.15} & 5.57 & 46.7 & 16.47 & \underline{13.32} & \underline{4.63} & $\approx$\underline{2000} & $\approx$\underline{60}\\
KoMA$^{*}$~\citep{jiang2024koma} & 73.3 & 23.00 & \textbf{13.84} & 5.44 & 66.7 & 22.63 & \textbf{15.83} & 4.66 & 60.0 & 20.37 & \textbf{14.89} & 4.43 & $\approx$5800 & $\approx$200\\
RESPOND~\citep{chen2026respond} & \underline{86.7} & \underline{26.37} & 12.29 & \underline{6.44} & \underline{83.3} & \underline{26.46} & 12.94 & \textbf{6.18} & \underline{76.7} & \underline{24.63} & 10.61 & \textbf{5.70} & $\approx$\textbf{1000} & $\approx$\textbf{50}\\
\midrule
\settingrow
\textbf{\xmind{} (ours)} & \textbf{93.3} & \textbf{28.13} & 10.74 & \textbf{7.19} & \textbf{90.0} & \textbf{27.53} & 11.65 & \underline{5.62} & \textbf{83.3} & \textbf{25.77} & 10.98 & 4.44 & $\approx$3100 & $\approx$180\\
\bottomrule
\end{tabular}%
}

\par\vspace{2pt}
\noindent
\begin{minipage}{\textwidth}
    \raggedright
    \footnotesize
    $^{*}$Single-agent variant. Each method retains its own memory-construction mechanism, while all other experimental settings are kept identical across methods.
\end{minipage}
\end{table*}

Retrieval starts when a surrounding vehicle reaches risk level 1 or higher. Candidate capsules must match the current scenario type and the direction of the highest-risk vehicle. Layout similarity $s_{\rho}(f_t,m_i)$ compares directional risk distributions, while text similarity $s_{\mathrm{text}}(f_t,m_i)$ measures semantic similarity between the scenario descriptions of $f_t$ and the capsule keyframe $f_{\tau_i^{*}}$. The retrieval score combines both:
\begin{equation}
s(f_t,m_i)
=
w\,s_{\rho}(f_t,m_i)
+
(1-w)s_{\mathrm{text}}(f_t,m_i),
\label{eq:memory_retrieval_score}
\end{equation}
where $w$ is the weight assigned to layout similarity.

On straight road segments where left and right symmetry is assumed, retrieval considers both the original and mirrored forms of each capsule, which is used to match left–right reversed risk patterns. The capsule with the highest score provides memory evidence $B_t$, including action classes, the preferred action, risk, and reflection. If no capsule is matched or its preferred action agrees with the initial action, the initial action is retained. Otherwise, the model thinks twice using $P_t$, $B_t$, and $W_t$:
\begin{equation}
a_t
=
\begin{cases}
a_t^{(1)},
& \text{no match or action agrees},\\
\pi_{\theta}(P_t,B_t,W_t),
& \text{otherwise}.
\end{cases}
\label{eq:final_online_action}
\end{equation}

The initial decision draws on the LLM's existing knowledge and its reasoning about current traffic. Scenario matching and the disagreement condition control when episodic evidence is introduced, helping reduce interference from weakly related memories. When no match is found or the action agrees, the model skips the second decision call, avoiding additional input tokens and inference cost.

\section{Experiments}
\label{sec:experiments}

The evaluation follows three questions: (Q1) Does the memory improve safety and efficiency over relevant baselines? (Q2) Does accumulation improve decisions on unseen real-data risk events? (Q3) Which mechanisms explain the difference? 

\subsection{Experimental Setup}
We evaluate \xmind{} in Highway-env~\citep{highway-env}, whose discrete meta-action space is
$\mathcal{A}=\{\texttt{LANE\_LEFT},\allowbreak
\texttt{IDLE},\allowbreak
\texttt{LANE\_RIGHT},\allowbreak
\texttt{FASTER},\allowbreak
\texttt{SLOWER}\}$ (Section~\ref{subsec:risk_wm}). The main experiments use three configurations with different traffic complexity: four lanes with density 2.5, five lanes with density 2.5, and five lanes with density 3.0. Each configuration is evaluated over 3 rounds of 30 episodes with different random seeds. Particularly, Q1 and Q3 use different test seeds to broaden the coverage of randomly generated traffic, so their results are not directly comparable. Each episode lasts 30 decision steps and succeeds if the ego vehicle completes all steps without collision. All experiments use GPT-6-Astra as the LLM backbone without fine-tuning.

We report success rate (SR), success steps (SS), mean speed $\bar v$, mean minimum TTC $\overline{\mathrm{TTC}}_{\min}$, and token usage per decision step. SR is the proportion of successful episodes, and SS is the mean number of executed decision steps per episode. $\bar v$ is averaged within each episode and then across episodes. $\overline{\mathrm{TTC}}_{\min}$ averages the minimum TTC over successful episodes. Input and output token counts are estimated per decision step.

\subsection{Q1: Baseline Comparison}

We compare \xmind{} with DiLu~\citep{wen2024dilu}, the single-agent variant of KoMA~\citep{jiang2024koma}, and RESPOND~\citep{chen2026respond}. Each method builds memory with its reflection mechanism on shared seeds excluded from testing, and the memory is fixed during evaluation.

\noindent\textbf{Safety metrics.} 
Table~\ref{tab:highway} summarizes the comparison. \xmind{} achieves success rates of 93.3\%, 90.0\%, and 83.3\%, ranking first in all three configurations. Relative to RESPOND, its success rate is higher by approximately 6.6 to 6.7 percentage points, and its mean success steps increase across configurations. With five lanes and a traffic density of 3.0, \xmind{} exceeds DiLu and KoMA in success rate by 36.6 and 23.3 percentage points, respectively. Within the fixed horizon of 30 decision steps, \xmind{} completes a greater proportion of episodes without collision, with mean episode lengths closer to the horizon limit.

With five lanes, increasing traffic density from 2.5 to 3.0 reduces the mean minimum TTC for all methods, indicating tighter temporal margins in denser traffic. \xmind{} records mean minimum TTC values of 7.19, 5.62, and 4.44\,s across the three configurations. Its successful trajectories therefore retain temporal margins of several seconds on average at their most critical approach points.

\noindent\textbf{Efficiency and cost.} 
Across the three traffic configurations, \xmind{}'s mean speed ranges from 10.74 to 11.65\,m/s, generally below DiLu and KoMA and close to RESPOND. Although DiLu and KoMA achieve higher mean speeds, both have lower success rates and success steps than \xmind{}. \xmind{} does not lead in speed across all configurations, but it completes more episodes without collision and executes more decision steps on average. Compared with RESPOND, which also incorporates risk information, \xmind{} achieves better episode completion at similar speed levels. These results illustrate the tradeoff between completion without collision and driving speed.

For token usage, \xmind{} uses approximately 3,100 input tokens and 180 output tokens per decision step. \xmind{} uses fewer tokens than KoMA, but its input and output token usage remains higher than that of DiLu and RESPOND. Part of this additional overhead comes from the second decision call triggered by disagreement. When the initial action differs from the preferred action in the first matched capsule, \xmind{} calls the LLM again with memory evidence, increasing both input and output token usage for that decision step.

\subsection{Q2: Experiments with Real Driving Data}
\label{subsec:krisk_evolution}

To test whether accumulated experience improves decisions on unseen events, we use the highD~\citep{krajewski2018highd}, NGSIM~\citep{fhwa2016ngsim}, and extreme subsets of K-Risk~\citep{huang2026knowledge}. The extreme subset includes events from ExpresswayA and FreewayB in CitySim~\citep{zheng2024citysim}. After standardizing trajectory units and direction conventions, we select three types of interaction: sudden braking by the lead vehicle and vehicles cutting in from the left or right. Selection considers braking and lane changes, time headway (THW), TTC, and the ego vehicle's response to retain interactions relevant to ego safety. Duplicate events are removed, and the resulting event counts are summarized in Table~\ref{tab:krisk_events}.

\begin{table}[t]
\centering
\caption{Counts of selected events across the K-Risk subsets.}
\label{tab:krisk_events}
\normalsize
\setlength{\tabcolsep}{4pt}

\begin{tabularx}{\columnwidth}{l*{3}{>{\centering\arraybackslash}X}}
\toprule
Event type & highD~\citep{krajewski2018highd} & NGSIM~\citep{fhwa2016ngsim} & extreme~\citep{zheng2024citysim}$^{*}$\\
\midrule
Hard braking & 42 & 35 & 38\\
Left cut in  & 23 & 30 & -\\
Right cut in & 21 & 42 & -\\
\midrule
\settingrow
\textbf{Total}
& \textbf{86} & \textbf{107} & \textbf{38}\\
\bottomrule
\end{tabularx}

\par\vspace{2pt}
\noindent
\begin{minipage}{\columnwidth}
    \raggedright
    \footnotesize
    $^{*}$Events from ExpresswayA and FreewayB in CitySim.
\end{minipage}
\end{table}

\noindent\textbf{Evaluation protocol.}
We compare Human, NoMem, and 30Mem at identical decision frames under the same counterfactual verification conditions. Human denotes the action extracted from the recorded trajectory. NoMem retains working memory and disables episodic memory retrieval. The 30Mem setting uses 30 risk capsules, which are constructed exclusively from simulator episodes separate from the K-Risk evaluation events, with memory updates disabled during evaluation.

\begin{figure*}[t]
    \centering
    \includegraphics[width=\textwidth]{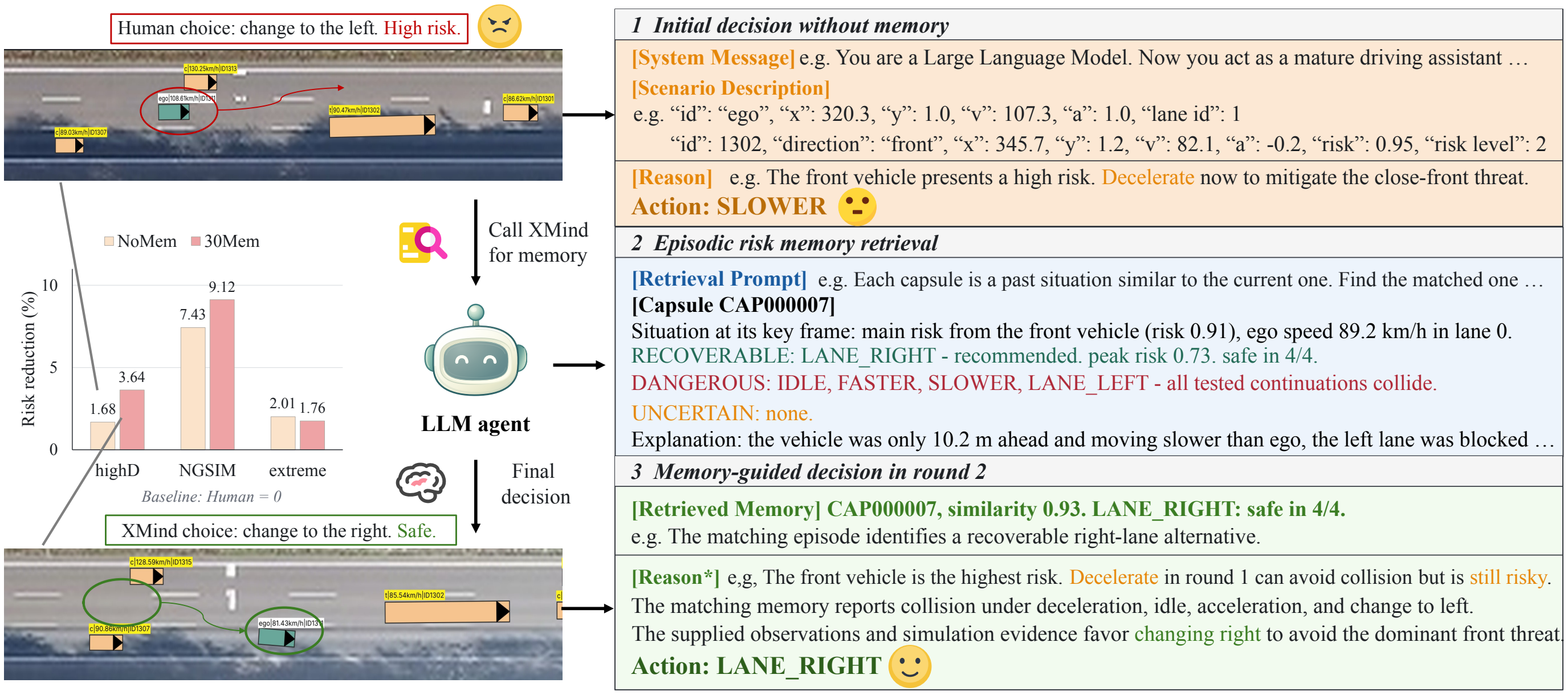}
    \caption{Risk reduction and memory-guided decision revision on K-Risk.
    The bar chart shows mean peak risk reduction relative to Human
    at paired critical frames (higher is better).
    The scenario contrasts the recorded Human left lane change
    with \xmind{}'s right lane change. The decision trace shows an initial
    \act{SLOWER} action, retrieval of a verified capsule,
    and a revised \act{LANE\_RIGHT} decision.
    Prompt and reasoning excerpts are illustrative.}
    \label{fig:krisk_risk_comparison}
\end{figure*}

\noindent\textbf{Risk reduction.}
The bar chart in Fig.~\ref{fig:krisk_risk_comparison} shows that NoMem reduces mean peak risk relative to Human by 1.68\% on highD, 7.43\% on NGSIM, and 2.01\% on extreme, while 30Mem achieves reductions of 3.64\%, 9.12\%, and 1.76\%, respectively. Relative to the common Human baseline, the reduction achieved by 30Mem exceeds that of NoMem by 1.96 percentage points on highD and 1.69 percentage points on NGSIM.

For a complementary safety comparison, we classify each method's outcome against Human as safer, similar, or riskier. The comparison first uses the unrecoverability fraction. When the fractions are equal, it uses mean peak risk with an absolute tolerance of 0.05. Table~\ref{tab:krisk_paired_safety} includes all valid paired decision frames. It shows that 30Mem is classified as safer than Human more often than riskier on all three datasets. Compared with NoMem, 30Mem increases the share of safer decisions on highD and NGSIM. The riskier share decreases on highD and rises slightly on NGSIM. On extreme, however, retrieved memory prompts two lane changes and one switch from \act{SLOWER} to \act{IDLE}, increasing the mean peak risk slightly above that of NoMem.

\begin{table}[t]
\centering
\caption{Pairwise safety comparison against Human (\%).}
\label{tab:krisk_paired_safety}
\small
\setlength{\tabcolsep}{4pt}
\renewcommand{\arraystretch}{1.08}

\begin{tabularx}{\columnwidth}{
    @{}>{\raggedright\arraybackslash}Xcccc@{}
}
\toprule
Dataset & Method & Safer & Similar & Riskier\\
\midrule
\multirow{2}{*}{\shortstack[l]{highD\\($n=319$)}}
& NoMem & 15.36 & 77.12 & 7.52\\
& 30Mem & 16.61 & 77.43 & 5.96\\
\midrule
\multirow{2}{*}{\shortstack[l]{NGSIM\\($n=468$)}}
& NoMem & 32.05 & 54.70 & 13.25\\
& 30Mem & 35.26 & 51.28 & 13.46\\
\midrule
\multirow{2}{*}{\shortstack[l]{extreme\\($n=138$)}}
& NoMem & 21.01 & 68.84 & 10.15\\
& 30Mem & 21.01 & 66.67 & 12.32\\
\bottomrule
\end{tabularx}

\par\vspace{2pt}
\noindent
\begin{minipage}{\columnwidth}
    \raggedright
    \footnotesize
    $n$ denotes valid paired decision frames.
    Similar outcomes do not require identical actions.
\end{minipage}
\end{table}

\noindent\textbf{Case study.}
The scenario and decision trace in Fig.~\ref{fig:krisk_risk_comparison} illustrate how retrieved evidence informs reconsideration. Human changes to the left in the recorded scenario, whereas the agent's initial decision without episodic evidence is \act{SLOWER} in response to a front vehicle with risk 0.95. Retrieval returns capsule \texttt{CAP000007} with high similarity, whose stored keyframe also contains a dominant front threat, with risk 0.91. In that stored event, \act{LANE\_RIGHT} admits a collision-free continuation in all four tested traffic conditions, with a reported peak risk of 0.73, whereas all tested continuations collide for \act{IDLE}, \act{FASTER}, \act{SLOWER}, and \act{LANE\_LEFT}. The capsule's preferred action disagrees with the initial \act{SLOWER} decision, triggering a second decision call. Using the retrieved evidence together with current observations, the agent selects \act{LANE\_RIGHT}. This example illustrates how counterfactual evidence from a related past event supports action revision without updating the LLM parameters.

\begin{table}[t]
\centering
\begin{minipage}{0.65\columnwidth}
\caption{Working memory window.}
\label{tab:working_memory_window}
\centering
\small
\setlength{\tabcolsep}{3pt}
\renewcommand{\arraystretch}{1.0}

\begin{tabularx}{\linewidth}{
  @{}
  >{\centering\arraybackslash}X
  >{\centering\arraybackslash}X
  >{\centering\arraybackslash}X
  @{}
}
\toprule
Window $N$ & SR (\%)$\uparrow$ & SS$\uparrow$ \\
\midrule
1 & 73.3 & 23.53 \\
3 & \textbf{80.0} & 25.27 \\
5 & \textbf{80.0} & \textbf{25.67} \\
\bottomrule
\end{tabularx}
\end{minipage}
\end{table}

\subsection{Q3: Ablation Study}
\label{subsec:ablation}

\begin{table*}[t]
\centering
\caption{Ablation study with five lanes and a traffic density
of 3.0. Text and Risk denote the two retrieval methods.}
\label{tab:component_ablation}
\small
\setlength{\tabcolsep}{4pt}
\renewcommand{\arraystretch}{1.12}

\begin{tabularx}{\textwidth}{
  >{\raggedright\arraybackslash}p{0.28\textwidth}
  *{4}{>{\centering\arraybackslash}X}
  >{\centering\arraybackslash}p{0.12\textwidth}
  *{2}{>{\centering\arraybackslash}X}
}
\toprule
\multirow{2}{*}{Method}
& Working
& Episodic
& \multicolumn{2}{c}{Retrieval method}
& Counterfactual
& \multirow{2}{*}{SR (\%)$\uparrow$}
& \multirow{2}{*}{SS$\uparrow$} \\[-2pt]
\cmidrule(lr){4-5}
\noalign{\vskip -1pt}
& memory
& memory
& Text
& Risk
& verification
& & \\
\midrule
Without memory
& \abno & \abno & \abno & \abno & \abno
& 50.0 & 17.03 \\

Working memory only
& \abyes & \abno & \abno & \abno & \abno
& 66.7 & 21.40 \\

Text retrieval
& \abyes & \abyes & \abyes & \abno & \abyes
& 73.3 & 23.03 \\

Risk retrieval
& \abyes & \abyes & \abno & \abyes & \abyes
& 80.0 & 24.83 \\

w/o counterfactual verification
& \abyes & \abyes & \abyes & \abyes & \abno
& 78.9 & 24.54 \\
\midrule
\settingrow
\xmind{} (different scenarios)
& \abyes & \abyes & \abyes & \abyes & \abyes
& \underline{83.3} & \underline{25.90} \\
\settingrow
\textbf{\xmind{} (matched scenarios)}
& \abyes & \abyes & \abyes & \abyes & \abyes
& \textbf{90.0} & \textbf{27.37} \\
\bottomrule
\end{tabularx}
\end{table*}

In this section, we compare working memory window sizes and conduct ablation studies to assess how memory components, retrieval methods, and counterfactual verification contribute to driving safety.

\noindent\textbf{Working memory window.}
We compare working memory windows of $N\in\{1,3,5\}$ decision frames on 15 seeds excluded from final evaluation. By Equation~\eqref{eq:working_memory}, the window includes the current frame, so $N=1$ provides only $f_t$. All settings share the same episodic memory $\mathcal M_E$, with updates disabled during evaluation.

Table~\ref{tab:working_memory_window} reports SR and SS. Both $N=3$ and $N=5$ reach an SR of $80.0\%$, versus $73.3\%$ for $N=1$, suggesting that short-term temporal context improves decision reliability. Although the five-frame window slightly increases SS from $25.27$ to $25.67$, it does not further improve SR. We therefore select $N=3$ as a better trade-off between temporal context and token cost.

\noindent\textbf{Component ablations.}
We conduct the component ablations under the configuration of five lanes and a traffic density of 3.0. We compare variants without working memory or counterfactual verification, together with an ablation of the retrieval methods. The variant without working memory receives only the current frame description. Text retrieval and risk retrieval rank capsules using $s_{\mathrm{text}}(f_t,m_i)$ and $s_{\rho}(f_t,m_i)$, respectively. Without counterfactual verification, the LLM generates the preferred action $a_i^{*}$ through reflection on the observed episode, with the remaining components unchanged. We also compare experience collected from different scenarios with experience from scenarios matching the evaluation setting. All other variants with episodic memory use different scenario experience and the number of capsules remains the same.

Table~\ref{tab:component_ablation} shows that working memory increases SR from $50.0\%$ to $66.7\%$ and SS from $17.03$ to $21.40$, indicating that short-term temporal context alone already improves decision reliability. The full method achieves an SR of $83.3\%$, compared with $73.3\%$ for text retrieval and $80.0\%$ for risk retrieval. This comparison supports combining scenario semantics with spatial risk information to retrieve more applicable experience. Removing counterfactual verification reduces SR from $83.3\%$ to $78.9\%$ and SS from $25.90$ to $24.54$, suggesting that reflection without explicit outcome verification can produce less reliable action recommendations. The source of experience also affects performance. Using matched scenario experience increases SR from $83.3\%$ to $90.0\%$ and SS from $25.90$ to $27.37$, showing that scene-relevant episodic memory provides stronger support for subsequent decisions.

\section{Conclusion}
We presented \xmind{}, a cross-memory reasoning framework for LLM-based autonomous driving. It combines working memory with counterfactually verified episodic risk memory to support ``Think Twice'' decision-making. At critical states, \xmind{} evaluates alternative actions and records whether they are recoverable, dangerous, or uncertain. When retrieved evidence disagrees with the initial decision, the agent reconsiders it with this evidence. In closed-loop highway simulation, \xmind{} achieves the highest success rate in all three configurations without updating the LLM. Ablations confirm the benefits of working memory, risk--text retrieval, and counterfactual verification. On recorded K-Risk events, simulator-derived experience reduces decision risk on highD and NGSIM, although its benefit remains limited on the extreme subset. Overall, \xmind{} shows that verified past experience can complement current traffic context for more reliable LLM-based driving. Future work will extend memory coverage to longer horizons and more diverse traffic scenarios.

\section{Acknowledgments}

This work was supported by the National Research Foundation of Korea (NRF) grant funded by the Korean government (MSIT) under the project “Development of Risk-Enhanced Continual Learning for Embodied Intelligence in Long-Tail Environments” (Grant No. RS-2026-25595451) and by the Korea Institute of Science and Technology Information (KISTI) R\&D program through the joint research project “Development of the Next-Generation Integrated Wired/Wireless Communication Gateway (X-Gateway).”

\bibliographystyle{IEEEtran}
\bibliography{refs}
\end{document}